\pdfoutput=1
\documentclass[11pt]{article}
\usepackage[final]{acl}
\usepackage{hyperref}

\usepackage{booktabs}  % 줄 꾸미기용
\usepackage{pgfplots}
\pgfplotsset{compat=1.18}
\usepackage{xcolor}
\usepackage{subcaption}
\usepackage{geometry}
\usepackage{hhline}
\usepackage{colortbl} 
\usepackage{xcolor} 
\usepackage{times}
\usepackage{latexsym}
\usepackage{amsmath}

\usepackage{algorithm}
\usepackage{algorithmic}
\usepackage{multirow}
\usepackage{tabularx}
\usepackage{array}  
\usepackage{caption}
\usepackage[T1]{fontenc}
\usepackage[utf8]{inputenc}

\usepackage{microtype}

\usepackage{inconsolata}

\usepackage{graphicx}
\usepackage{kotex}
\usepackage{enumitem}

\title{CCQA: Generating Question from Solution Can Improve Inference-Time Reasoning in SLMs}

\author{
  Jin Young Kim\textsuperscript{1} \quad
  Ji Won Yoon\textsuperscript{1}\thanks{Corresponding author.}\\
  \textsuperscript{1}Department of Artificial Intelligence, Chung-Ang University\\
  \texttt{\{wlsdud338, jiwonyoon\}@cau.ac.kr}
}
\begin{document}
\maketitle  
\begin{abstract}

Recently, inference-time reasoning strategies have further improved the accuracy of large language models (LLMs), but their effectiveness on smaller models remains unclear.
Based on the observation that conventional approaches often fail to improve performance in this context, we propose \textbf{C}ycle-\textbf{C}onsistency in \textbf{Q}uestion \textbf{A}nswering (CCQA), a novel reasoning method that can be effectively applied to SLMs. Inspired by cycle consistency, CCQA generates a question from each reasoning path and answer, evaluates each by its similarity to the original question, and then selects the candidate solution with the highest similarity score as the final response.
Since conventional SLMs struggle to generate accurate questions from their own reasoning paths and answers, we employ a lightweight Flan-T5 model specialized for question generation to support this process efficiently.
From the experimental results, it is verified that CCQA consistently outperforms existing state-of-the-art (SOTA) methods across eight models on mathematical and commonsense reasoning benchmarks. Furthermore, our method establishes a new practical baseline for efficient reasoning in SLMs. Source code can be found at \url{https://github.com/scai-research/ccqa_official}.
\end{abstract}

\section{Introduction}
Recent advancements in large language models (LLMs) have yielded remarkable performance across a wide range of tasks, including machine translation \cite{machine_translation1,machine_translation2}, code generation \cite{code_generation1,code_generation2}, sentiment analysis \cite{sentiment_analysis1,sentiment_analysis2}, and reasoning \cite{math_reasoning1,commonsense_reasoning1}. On top of that, inference-time reasoning strategies, such as chain-of-thought (CoT) \cite{CoT}, self-consistency (SC) \cite{self_consistency}, and self-correction \cite{self_correction}, can produce more reliable outputs and further improve model accuracy, albeit at the cost of additional test-time computation \cite{CoT,self_consistency,self_correction}.

While prior studies have clearly demonstrated the effectiveness of these reasoning techniques for large-scale models \cite{self_consistency, self_correction, self_refinement}, their applicability to small language models (SLMs) has yet to be fully explored. This motivates us to empirically investigate whether such reasoning strategies remain effective when applied to SLMs, and our observations indicate that they often lead to performance degradation in this setting, which will be discussed in Section \ref{main_result}.

The performance degradation observed in SLMs can be attributed to two main factors. First, smaller models could struggle to understand complex inputs and fail to follow instructions \cite{scailing_parameter,in-context_length,in-context_leraning_slm_limitation,in-context_leraning_slm_limitation2}. However, recent self-feedback methods, such as self-correction \cite{self_correction}, self-refinement \cite{self_refinement}, and universal self-consistency (USC) \cite{USC}, operate under the assumption that the model is capable of comprehending lengthy and complex inputs to generate appropriate feedback. This mismatch between the model’s capacity and the underlying assumption often leads to suboptimal or even misleading outputs in the context of SLMs. 
Second, voting-based approaches such as SC \cite{self_consistency} rely on a majority vote across multiple generated answers. This strategy becomes less effective when SLMs produce highly inconsistent outputs \cite{inconsistency}. 
In such cases where generated answers exhibit high variance without a clearly dominant response, majority voting fails to produce a reliable consensus and offers no meaningful advantage over random choice.
This occurs because SC selects the final answer solely based on frequency, without evaluating the quality of reasoning paths.

To address these limitations, we propose a novel reasoning method that can be effectively applied to SLMs, called \textbf{C}ycle-\textbf{C}onsistency in \textbf{Q}uestion \textbf{A}nswering (CCQA). 
Inspired by the principle of cycle consistency \cite{Cycle_consistency}, we construct a cycle between the original question, the solution produced by the SLM, and the question generated from that solution. Here, the solution includes a reasoning path and its corresponding answer. We believe that if the reasoning path and answer are correct, the regenerated question should be highly similar to the original input question. In the proposed framework, the SLM first receives the question as input and produces multiple candidate solutions. When there is no dominant response during majority voting, CCQA generates a new question from each candidate and measures its similarity to the original; a higher similarity score indicates that the solution is more likely correct.
The candidate solution whose generated question most closely matches the original is selected as the final response, without requiring the model to process any additional complex input.
% This approach addresses SLMs' limitations in processing long feedback prompts. 
% We only need to select the answer whose generated question best matches the original question, without requiring the model to process additional complex input.
Moreover, we fine-tune a lightweight Flan-T5-base \cite{T5} model to generate questions from candidate solutions.
This is because conventional SLMs typically struggle to generate questions from their reasoning paths and answers.
We confirm that our fine-tuned Flan-T5 is both efficient and excels at producing high-quality questions.

Extensive experiments are conducted on six reasoning benchmarks, including four mathematical and four commonsense tasks. Our evaluation uses eight SLMs ranging from 135M to 3B parameters, including Llama3.2 \cite{llama}, SmolLM2 \cite{smollm}, and Qwen2.5 \cite{qwen}. From the experimental results, it is confirmed that CCQA consistently outperforms current state-of-the-art (SOTA) reasoning methods across most SLMs and benchmarks. Notably, CCQA with Llama3.2-3B on GSM8K achieves 69.60\% accuracy compared to USC’s 53.83\%. On CommonSenseQA with Llama-1B, it attains 38.74\% versus USC’s 33.99\%, demonstrating its effectiveness in enhancing SLM reasoning capabilities.

Our main contributions are summarized as follows:
\begin{itemize}
    \item Our paper introduces a novel inference-time reasoning technique for SLMs, namely CCQA, that evaluates the quality of each reasoning path and its answer by regenerating a question and measuring its similarity to the original. \textit{To the best of our knowledge, this is the first attempt to investigate the inference-time reasoning capabilities of SLMs and to improve them.}
    % \item We introduce CCQA, a novel approach of cycle consistency principles to reasoning tasks, generating a question-solution-question framework that effectively evaluates reasoning quality without requiring complex feedback mechanisms.
    % \item We theoretically and experimentally demonstrated that existing methods designed for LLMs significantly degrade in performance when applied to SLMs, clearly identifying the reasons for this phenomenon.
    \item We leverage a lightweight Flan-T5 model to generate questions from candidate solutions. Compared to conventional SLMs, our fine-tuned Flan-T5 is computationally efficient and produces higher-quality questions.
    \item Our extensive experiments across diverse benchmarks and SLMs demonstrate that CCQA consistently outperforms SOTA reasoning methods, substantially improving reasoning capabilities of SLMs.
\end{itemize}
\begin{figure*}
    \centering
    \includegraphics[width=\textwidth]{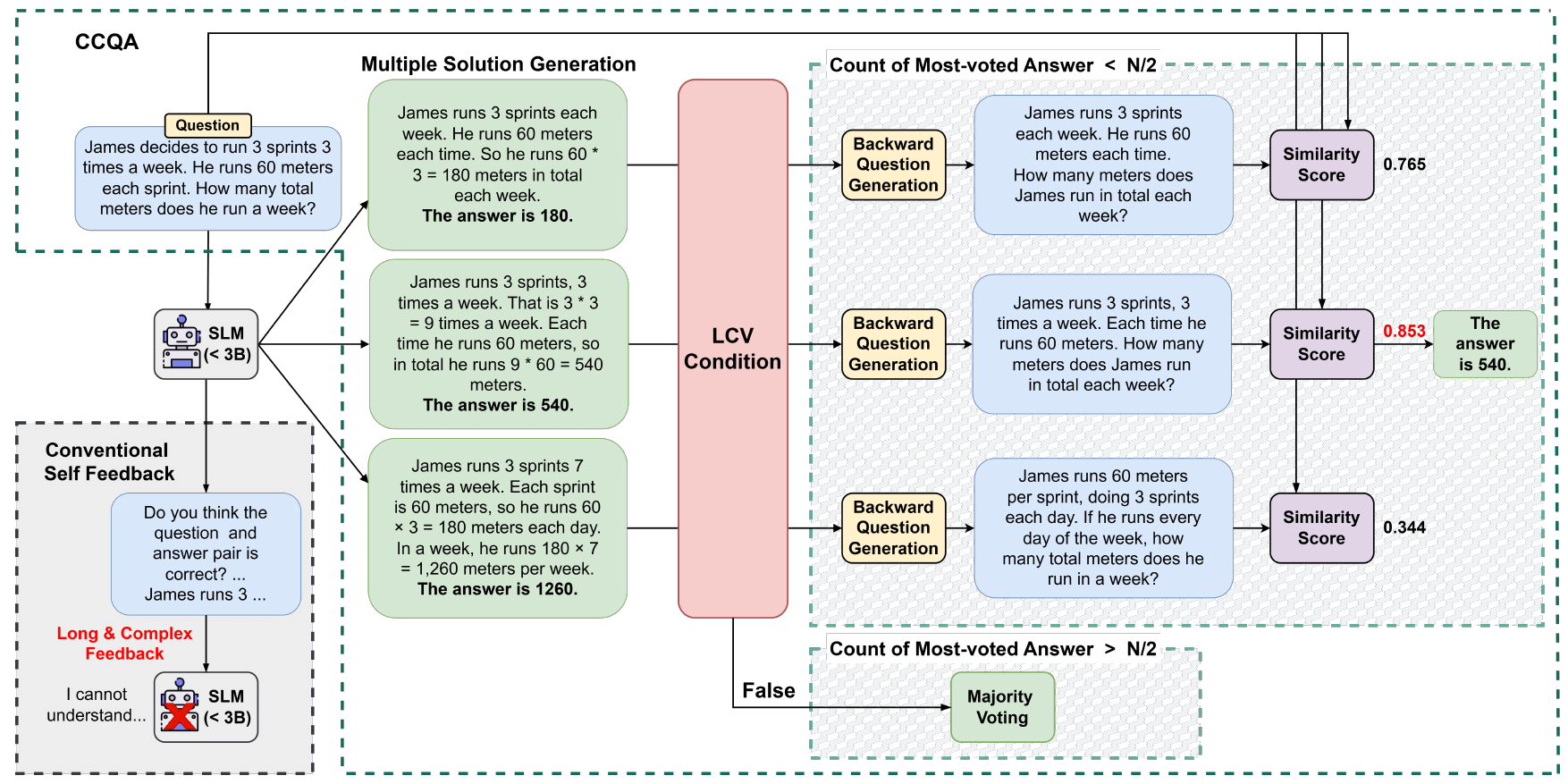}
    \caption{Overall process of the CCQA. (1) CCQA receives a question as input and generates $N$ solutions. (2) It checks for the LCV condition; if the LCV condition is met (i.e., when the model’s answers are inconsistent with no clear majority), it regenerates questions from the answers, otherwise it performs majority voting to select the final answer. (3) Under the LCV condition, it compares the generated questions with the original question to assign similarity scores. (4) The solution corresponding to the question with the highest similarity score is selected as the final answer.}
    \label{fig:ccqa}
\end{figure*}
\section{Related Work}\label{related_work}
\subsection{Reasoning Methods for LLMs in Test-time}
Reasoning remains one of the most challenging tasks for language models, involving complex problem-solving such as arithmetic and commonsense reasoning. Various approaches have been developed to enhance reasoning performance. CoT \cite{CoT} induces models to describe problem-solving steps clearly, with extensions like least-to-most \cite{least2most} and tree of thought \cite{ToT} exploring more diverse reasoning paths. Self-feedback methods, including self-correction \cite{self_correction} and self-refinement \cite{self_refinement}, enable models to improve outputs using their own responses, though these typically require processing extensive input contexts. Aggregation techniques such as SC \cite{self_consistency} employ majority voting across multiple samples, while USC \cite{USC} generates responses by considering all previous outputs. However, these methods rely on assumptions about model capacity that become problematic for SLMs with inconsistent outputs or limited ability to process complex prompts.

\subsection{Cycle Consistency in Generative Models}
Cycle consistency has been widely used as an effective training and evaluation paradigm across various domains. Initially introduced in computer vision for tasks such as image-to-image translation \cite{CGAN} and 3D reconstruction \cite{3D-cycle-consistency}, the concept leverages the principle that transformations should be reversible — if data is transformed from domain A to domain B and back to domain A, the result should closely match the original input. This principle has been extended to natural language processing, including machine translation \cite{MT-mong,MT-french}, where back translation serves as a form of cycle consistency to improve translation quality. Recent work has also explored cycle consistency for evaluating text generation quality \cite{QA-cycle-consistency} and ensuring factual consistency in summarization \cite{cycle-consistency-summarization}. While cycle consistency has been used to assess generation quality in various domains, applying it to SLM's reasoning quality offers a promising new direction.

\section{Proposed Method}
\subsection{Motivation}
SLMs have gained increasing attention \cite{on_device_survey1,need_for_on_device}, but when applied to reasoning tasks, they underperform mainly due to two limitations. First, SLMs struggle with processing long and complex inputs (>1K tokens) due to their weak in-context learning abilities \cite{lost_in_middle,in-context_length,in-context_leraning_slm_limitation}. Second, they often generate inconsistent outputs that are highly varied \cite{inconsistency}. Consequently, both self-feedback mechanisms (e.g., self-correction, self-refinement) and voting mechanisms (e.g., SC, USC) show limited effectiveness with SLMs, as they require either strong input processing capabilities or output consistency. Despite these limitations, reasoning approaches specifically designed for SLMs remain largely unexplored, highlighting the need for tailored methodologies for SLMs. 
Based on these observations, we derive two key requirements for effective SLM reasoning: (1) avoid lengthy feedback or correction prompts, and (2) reliably identify high-quality reasoning despite inconsistent outputs.
% Based on these observations, we identify two key requirements for an effective SLM reasoning approach: (1) it should avoid requiring the model to process lengthy prompts for feedback or correction, and (2) it should provide a reliable mechanism to identify high-quality reasoning even when faced with inconsistent outputs. 

\subsection{CCQA}
The overall process of CCQA is illustrated in Figure \ref{fig:ccqa}. CCQA begins by generating $N$ independent solutions, including both reasoning paths (RPs\footnote{Reasoning paths refer to the step-by-step solution processes generated using CoT prompting, excluding the final answer.}) and answers, using CoT prompting. It then applies answer-only voting, as in the SC method, disregarding RPs. However, SLMs frequently produce extremely diverse answers to the same question, leading to unstable voting patterns where majority voting acts like random selection. In those cases, we employ a fine-tuned T5 model to regenerate a question from each solution (Section \ref{sec:bqg}) and measure both lexical and semantic similarity between each generated question and the original one. Finally, CCQA determines the final output by selecting the answer whose generated question has the highest similarity score
 (Section \ref{sec:sim}).

\subsection{Multiple Solution Generation and LCV Identification}
\label{sec:lcv}
When SLMs generate highly varied answers, additional verification becomes necessary. For instance, consider a case where a mathematical problem yields answers `18', `24', `27', and `35' with similar frequencies across multiple generated solutions. In this scenario, it is difficult to determine which answer is more reliable based on voting alone, as the method cannot evaluate the quality of the RPs. To address this problem, we define such situations with highly varied answers as \textbf{L}ow \textbf{C}onfidence \textbf{V}oting (LCV) conditions. An LCV condition is defined by the following condition:
\[
\text{LCV} = \bigl\{\,\max_j \text{freq}(A_j) < \lceil N/2 \rceil \bigr\}.
\]

Here, $\text{freq}(A_j)$ represents the frequency of the $j$-th unique answer, and $N$ is the total number of generated responses. In other words, LCV is defined as a situation where the frequency of the most voted answer does not reach a majority of the total number of responses. In our experiments on the GSM8K dataset, evaluating eight different models of various sizes (0.5B-3B), we found that on average, LCV occurred in 36.46\% of problems, and 80.85\% of answers selected by SC in these LCV cases were incorrect. This demonstrates that a simple majority voting method cannot sufficiently leverage the reasoning capabilities of SLMs. In these situations, a verification mechanism that directly considers RPs is needed rather than a majority voting approach. Therefore, we apply backward question generation and similarity measurement methods to directly evaluate the quality of each RP.

\subsection{Backward Question Generation}
\label{sec:bqg}
To evaluate RP quality in LCV situations, we leverage backward question generation. 
The backward question generation process is as follows: In LCV situations, each $RP_i$ (where $i$ indicates the index of the RP) is used as input to the fine-tuned T5 model to generate a question ($GQ_i$, where $i$ indicates the index of the GQ). To ensure more accurate question generation, we carefully select the appropriate model architecture and design a comprehensive training process. 
Among models of similar size, we choose the T5-base model due to its superior performance in text generation tasks. We also experiment with other small-sized models, but they do not perform well regardless of whether we apply fine-tuning or not. Furthermore, we utilize training sets from various mathematical and commonsense reasoning benchmarks, reverse the existing question-answer pairs to answer-question format for our task. Detailed hyperparameters and data preprocessing rules are presented in the Section \ref{setting}.
%==================================수도코드====================================================================
\begin{algorithm}[t]
\caption{CCQA}\label{alg:ccqa}
\begin{algorithmic}[1]
\REQUIRE Original Question $OQ$, parameter $\alpha$, $\beta$, sample count $N$, backward question generation BQG, reasoning path $RP_i$, answer $A_i$
\ENSURE Final answer $(RP_{final}, A_{final})$

\STATE Generate reasoning path and answers $\{(RP_i, A_i)\}_{i=1}^N$

\STATE Count frequency of each unique answer: freq($A_j$)
\STATE $j_{max} \leftarrow \arg\max_j \text{freq}(A_j)$

\IF{freq($A_{j_{max}}$) $\geq \lceil N/2 \rceil$}
    \STATE \textbf{return} ($RP_{j_{max}}$, $A_{j_{max}}$)
\ELSE
  \FOR{$i=1$ to $N$ do}
    \STATE $GQ_i \leftarrow$ BQG($RP_i$)
    \STATE $bleu_i \leftarrow$ BLEU($GQ_i$, OQ)
    \STATE $cos_i \leftarrow$ Cosine similarity($GQ_i$, OQ)
    \STATE $score_i \leftarrow \alpha \cdot bleu_i + \beta \cdot cos_i$
  \ENDFOR
  
  \STATE $best\_idx \leftarrow \arg\max_i score_i$
  \STATE return $(RP_{best\_idx}, A_{best\_idx})$
\ENDIF
\end{algorithmic}
\end{algorithm}
%=================================================================================================================
\subsection{Similarity-based Answer Selection}
\label{sec:sim}

After the backward question generation is completed, we need to compare the similarity between the generated questions ($GQ$) and the original question ($OQ$). We measured the similarity between each $GQ_i$ from each reasoning path $RP_i$ and the original question ($OQ$) using two complementary methods: BLEU \cite{bleu} and embedding-based cosine similarity \cite{sbert}. BLEU score for lexical overlap and embedding-based cosine similarity for semantic correspondence. BLEU score captures the lexical overlap and structural similarity by measuring n-gram matches between the generated and original questions. This helps identify how well the surface-level textual elements are preserved. For semantic similarity, we used cosine similarity of sentence embeddings generated by Sentence-BERT \cite{sbert}, which captures the overall meaning correspondence between the two questions beyond exact word matches.
These two measurements were combined using the following weighted sum:
\begin{align}
\text{score}(GQ_i, OQ) &= \alpha \cdot \text{BLEU}(GQ_i, OQ) \notag \\
&+ \beta \cdot \text{cosine}(GQ_i, OQ).
\end{align}
Here, BLEU is the BLEU score value, and cosine is the embedding-based cosine similarity score value. $\alpha$ and $\beta$ are weights that adjust the importance of each measurement. In our method, we set $\alpha$ to 0.4 and $\beta$ to 0.6. Detailed experiments for determining these weights are presented in Section \ref{similarity_metric}. The complete CCQA approach is formalized in Algorithm \ref{alg:ccqa}.

\section{Experimental Setup}
\label{setting}

\paragraph{Models.}
The specific models used in our experiments are Llama3.2-1B and Llama3.2-3B \cite{llama}, Qwen2.5-0.5B, Qwen2.5-1.5B, and Qwen2.5-3B \cite{qwen}, SmolLM2-135M and SmolLM2-360M \cite{smollm}.
Llama3.2 is a decoder-only language model with improved reasoning and instruction-following capabilities. We selected Llama3.2-1B and Llama3.2-3B variants to test performance on recent architectural designs.
Qwen2.5 is a transformer-based model known for its strong multilingual capabilities and performance on knowledge-intensive tasks. To assess how CCQA's scaling properties are affected by increasing model capacity, we utilized three different variants of this model.
SmolLM2 is a lightweight model optimized for efficiency with a specialized architecture for resource-constrained environments. We included SmolLM2-135M and SmolLM2-360M variants to test CCQA's applicability in on-device environments. Also we fine-tuned Flan-T5-base(258M) models to generate question, using learning rate of 2e-5, 3 epochs, and a batch size of 16.
\paragraph{Benchmarks.}
We evaluated CCQA on six standard reasoning benchmarks. For arithmetic reasoning, we utilized GSM8K \cite{gsm8k} with its multi-step grade school math problems (train: 7.47K, test: 1.32K), SVAMP \cite{svamp} offering varied math word problems (train: 700, test: 300), and Multi-Arith \cite{multiarith} for problems requiring multiple operations (train: 420, test: 180). For commonsense reasoning, we selected CSQA \cite{csqa} for its multiple-choice questions requiring world knowledge (train: 9.74K, val: 1.22K, test: 1.14K), StrategyQA \cite{strategyqa} which poses yes/no questions needing strategic inference (train: 1.6K, test: 687), and ARC-Challenge \cite{arc} (train: 1.12K, val: 299, test: 1.17K). If there was an answer field, we used the test dataset; if not, we used the dev dataset. Also, the datasets used for finetuning were the train sets of CSQA, StrategyQA, and GSM8K-main.
\paragraph{Implementation.}
We conducted experiments using the A6000 with 48GB. For generating model responses, we followed standard guidelines to set the temperature parameter for text generation \cite{temperature1,temperature2}. Specifically, we configured the temperature to 0.7 across all models when generate solutions. Additionally, based on previous research showing that top-p sampling provides more stable results for smaller models\cite{topp1,topp2}, we used top-p = 0.9 for decoding. \cite{topp1,topp2}. We conducted all experiments in a few-shot setting, utilizing demonstration examples derived from prior open-domain text generation studies \cite{CoT,self_consistency}. We also created simple prompts for T5 question generation. Our prompts and sample solutions are presented in Appendix \ref{app: a}, which shows the corresponding prompts used for question generation. Additionally, we converted question-answer pairs from the training sets of various reasoning benchmarks into answer-question pairs to fine-tune the Flan-T5 model. 
%-===================================table1========================================
\begin{table*}[t]
\centering
\renewcommand{\arraystretch}{1.1}
\setlength{\tabcolsep}{2.5pt}  % 열 간격 조정
\small  % 글씨 크기 줄이기
\resizebox{\textwidth}{!}{%
\begin{tabular}{l|cccc>{\columncolor{gray!15}}c|cccc>{\columncolor{gray!15}}c|cccc>{\columncolor{gray!15}}c}
\hline
\multirow{2}{*}{Model} & \multicolumn{5}{c|}{GSM8K} & \multicolumn{5}{c|}{MultiArith} & \multicolumn{5}{c}{SVAMP} \\
\hhline{~---------------}
 & Base & CoT & Self-Corr & SC & CCQA & Base & CoT & Self-Corr & SC & CCQA & Base & CoT & Self-Corr & SC & CCQA \\
\hline
Qwen-0.5B & 2.65 & 11.45 & 4.55 & \textbf{17.32} & \textbf{17.32} & 8.89 & 40.00 & 11.24 & 51.11 & \textbf{52.22} & 6.33 & 44.33 & 15.33 & 52.67 & \textbf{55.00} \\
Qwen-1.5B & 8.19 & 37.98 & 22.37 & 44.88 & \textbf{48.37} & 21.67 & 95.00 & 66.85 & \textbf{97.22} & \textbf{97.22} & 3.00 & 74.33 & 38.00 & 83.67 & \textbf{84.00} \\
Qwen-3B & 9.78 & 33.01 & 0.30 & 29.12 & \textbf{30.71} & 42.22 & 75.56 & 0.00 & 81.11 & \textbf{82.78} & 3.33 & 86.00 & 16.67 & 88.00 & \textbf{88.33} \\
Llama-1B & 2.05 & 25.32 & 17.89 & 35.78 & \textbf{39.20} & 8.89 & 70.22 & 25.84 & 85.00 & \textbf{86.11} & 2.66 & 52.33 & 40.00 & 58.67 & \textbf{59.00} \\
Llama3.2-3B & 1.59 & 49.81 & 4.85 & 69.31 & \textbf{69.60} & 21.11 & 80.58 & 8.63 & 93.89 & \textbf{98.89} & 4.66 & 79.00 & 27.33 & 85.00 & \textbf{86.00} \\
Falcon-1B & 5.76 & 32.52 & 0.08 & 40.94 & \textbf{42.61} & 7.22 & 79.21 & 63.33 & 91.67 & \textbf{92.78} & 3.33 & 44.00 & 20.67 & 51.33 & \textbf{52.33} \\
SmolLM2-135M & 1.90 & 2.35 & 0.00 & 1.97 & \textbf{2.88} & 0.56 & 0.00 & 0.00 & \textbf{3.33} & \textbf{3.33} & 5.00 & 6.00 & 0.00 & \textbf{8.33} & 7.67 \\
SmolLM2-360M & 2.65 & 6.60 & 0.00 & \textbf{8.79} & 8.72 & 1.67 & 7.78 & 0.00 & 24.44 & \textbf{25.56} & 1.66 & 15.33 & 0.00 & 24.33 & \textbf{27.00} \\
\hhline{================} 
\multirow{2}{*}{Model} & \multicolumn{5}{c|}{CommonSenseQA} & \multicolumn{5}{c|}{StrategyQA} & \multicolumn{5}{c}{ARC-Challenge} \\
\hhline{~---------------} 
 & Base & CoT & Self-Corr & SC & CCQA & Base & CoT & Self-Corr & SC & CCQA & Base & CoT & Self-Corr & SC & CCQA \\
\hline
Qwen-0.5B & 40.33 & 39.64 & 21.21 & 43.00 & \textbf{43.82} & 52.33 & 53.13 & 19.07 & \textbf{54.29} & \textbf{54.29} & 44.96 & 44.28 & 26.54 & 48.46 & \textbf{49.89}  \\
Qwen-1.5B & 57.56 & 62.74 & 7.53 & \textbf{66.58} & 66.34 & 51.38 & 55.02 & 12.23 & 52.55 & \textbf{55.17} & 66.19 & 71.33 & 24.66 & \textbf{75.09} & 74.40 \\
Qwen-3B & 65.26 & 65.11 & 70.27 & \textbf{70.52} & \textbf{70.52} & 54.15 & 51.97 & 51.53 & 52.98 & \textbf{55.31} & 76.19 & 79.95 & 26.56 & 84.90 & \textbf{84.98} \\
Llama-1B & 24.07 & 30.71 & 22.43 & 37.92 & \textbf{38.74} & 53.28 & 54.00 & 3.64 & 57.21 & \textbf{57.35} & 45.34 & 44.71 & 39.25 & 49.40 & \textbf{49.66} \\
Llama3.2-3B & 43.39 & 56.18 & 48.37 & 65.68 & \textbf{66.42} & 48.47 & 45.65 & \textbf{51.53} & 49.20 & 49.20 & 68.63 & 69.71 & 72.05 & \textbf{74.40} & 74.06 \\
Falcon-1B & 28.32 & 32.68 & 0.00 & 35.14 & \textbf{35.79} & 54.04 & 58.52 & 6.11 & 58.52 & \textbf{59.57} & 51.96 & 54.52 & 30.72 & 55.38 & \textbf{55.72} \\
SmolLM2-135M & 16.09 & 16.79 & 0.00 & 17.69 & \textbf{18.35} & 48.47 & 49.05 & 0.87 & \textbf{49.34} & \textbf{49.34} & 17.41 & 22.53 & 18.00 & 23.12 & \textbf{23.98} \\
SmolLM2-360M & 18.96 & 19.66 & 0.00 & 19.49 & \textbf{19.82} & 39.74 & 49.49 & 0.15 & 49.20 & \textbf{49.49} & 16.09 & 16.81 & \textbf{18.21} & 18.00 & 18.09 \\
\hline
\end{tabular}
}
\caption{Performance comparison including baseline and various inference-time techniques on arithmetic(GSM8K, Multi-Arith, SVAMP) and common-sense(CommonSenseQA, StrategyQA, ARC-Challenge) benchmarks, measured by accuracy(\%).
Base: baseline performance using greedy decoding, CoT: chain-of-thought prompting, Self-Corr: self-correction, SC: self-consistency, CCQA: proposed method.}
\label{tab:performance}
\end{table*}
%==========================table1 end===================================================

\section{Experimental Results} \label{main_result}
We evaluated CCQA on multiple benchmarks, comparing it with conventional reasoning methods, including CoT \cite{CoT}, self-correction \cite{self_correction}, SC \cite{self_correction}, and USC \cite{USC}.
As mentioned earlier, we focused on assessing the effectiveness of inference-time reasoning strategies for SLMs.

\subsection{Main Results}
\paragraph{Arithmetic Reasoning.}
The results for arithmetic reasoning are presented in Table \ref{tab:performance}.
Each SLM independently generated five solutions ($N=5$) per question.
Interestingly, conventional feedback-based methods, such as self-correction and USC, showed significant performance degradation. 
When applied to SLMs, these methods achieved markedly lower accuracy than chain-of-thought prompting and self-consistency, suggesting that feedback-dependent inference strategies may be suboptimal in the smaller model setting.
However, the results show that CCQA consistently achieves the highest accuracy across most configurations, outperforming all other methods on GSM8K, SVAMP, and MultiArith.
In particular, on the MultiArith benchmark, Llama3.2-3B with CCQA achieves a 5.00 \% performance improvement over SC, showed better accuracy than the other methods.
For SmolLM2, its limited capacity resulted in generally poor performance on the arithmetic reasoning benchmark. Nevertheless, CCQA still produced a measurable accuracy improvement.

\paragraph{Commonsense Reasoning.}
Experimental results for commonsense reasoning are also summarized in Table \ref{tab:performance}.
Across the three evaluated benchmarks, including CommonsenseQA, StrategyQA, and ARC-Challenge, CCQA outperformed competing methods across most models and benchmarks.
As with arithmetic reasoning, self-correction and USC also exhibited performance degradation in commonsense reasoning.
For SmolLM2, every method except CCQA and SC performed worse than CoT, with some approaches’ accuracy even falling to 0 \%. Among the techniques, only CCQA and SC maintained performance at or above CoT.
However, CCQA consistently delivered higher accuracy than SC.
This suggests that CCQA provides more consistent gains for SLMs across both arithmetic and commonsense reasoning tasks, whereas other inference‐time methods may sometimes underperform or show unstable results.

\subsection{CCQA Performs More Robustly Under LCV Condition}
\begin{table}[t]
\centering
\small
\begin{tabular}{lccccc}
\hline
\textbf{Benchmark }& LCV & SC$_{\text{LCV}}$ & CCQA$_{\text{LCV}}$ & $\Delta$ \\
\hline\hline
GSM8K        & 36.46  & 19.15 & \textbf{22.11} & \textbf{+2.96}  \\
CSQA       & 21.28 & 26.48 & \textbf{28.68} & \textbf{+2.20}  \\
StrategyQA & 5.79 & 36.88 & \textbf{48.03} & \textbf{+11.15} \\
SVAMP    & 36.46  & 19.15 & \textbf{21.20} & \textbf{+2.05} \\
\hline
\end{tabular}
\caption{Proportion of questions triggering the LCV condition and corresponding accuracy.
LCV (\%) = percentage of all samples under LCV condition (i.e., no clear majority);
SC$_{\text{LCV}}$ (\%) = accuracy of self-consistency on LCV samples;  
CCQA$_{\text{LCV}}$ (\%) = accuracy of the proposed method on the same samples;  
$\Delta$ (percentage-point gain) = CCQA$_{\text{LCV}}$ – SC$_{\text{LCV}}$.}
\label{tab:lcv_gain}
\end{table}

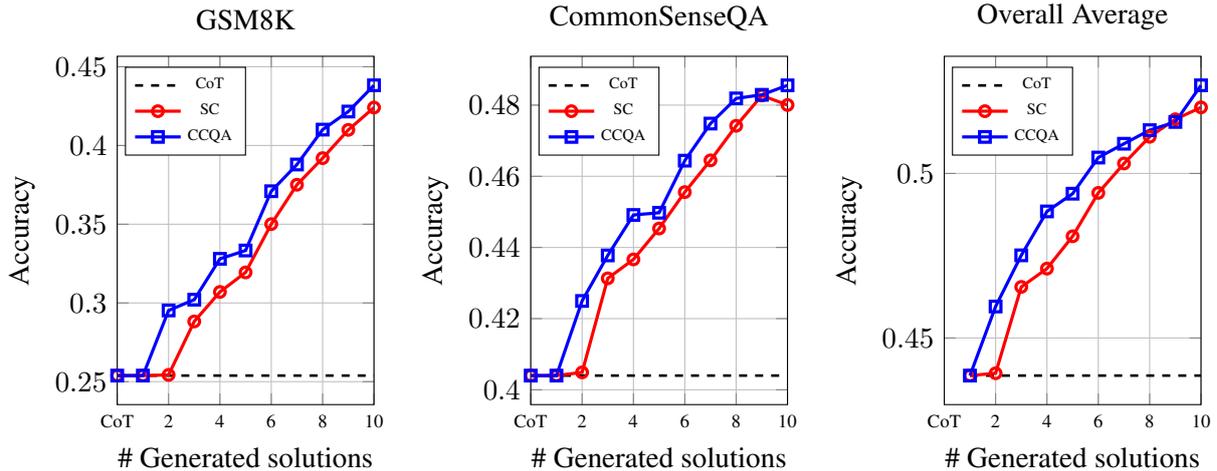
\begin{figure*}[t]
\centering

% ───── GSM8K ─────
\begin{subfigure}[t]{0.31\textwidth}
\centering
\begin{tikzpicture}
\begin{axis}[
    width=\linewidth,
    height=6.2cm,
    xlabel={\# Generated solutions},
    ylabel={Accuracy},
    title={GSM8K},
    xmin=0, xmax=10,
    xtick={0,2,4,6,8,10},
    xticklabels={CoT,2,4,6,8,10},
    x tick label style={font=\scriptsize},
    grid=both,
    legend pos=north west,
    legend style={font=\tiny},
]
\addplot[
    color=black,
    dashed,
    line width=1pt,
] table[x=k, y=CoT, col sep=comma] {graph/gsm8k_avg_sc_vs_ccqa.csv};
\addlegendentry{CoT}
\addplot[
    color=red,
    mark=o,
    mark size=2pt,
    line width=1.2pt,
] table[x=k, y=SC, col sep=comma] {graph/gsm8k_avg_sc_vs_ccqa.csv};
\addlegendentry{SC}

\addplot[
    color=blue,
    mark=square,
    mark size=2pt,
    line width=1.2pt,
] table[x=k, y=CCQA, col sep=comma] {graph/gsm8k_avg_sc_vs_ccqa.csv};
\addlegendentry{CCQA}
\end{axis}
\end{tikzpicture}
\end{subfigure}%
\hspace{1em}
% ───── CommonSenseQA ─────
\begin{subfigure}[t]{0.31\textwidth}
\centering
\begin{tikzpicture}
\begin{axis}[
    width=\linewidth,
    height=6.2cm,
    xlabel={\# Generated solutions},
    ylabel={Accuracy},
    title={CommonSenseQA},
    xmin=0, xmax=10,
    xtick={0,2,4,6,8,10},
    xticklabels={CoT,2,4,6,8,10},
    x tick label style={font=\scriptsize},
    grid=both,
    legend pos=north west,
    legend style={font=\tiny},
]
\addplot[
    color=black,
    dashed,
    line width=1pt,
] table[x=k, y=CoT, col sep=comma] {graph/commonSenseQA_avg_sc_vs_ccqa.csv};
\addlegendentry{CoT}

\addplot[
    color=red,
    mark=o,
    mark size=2pt,
    line width=1.2pt,
] table[x=k, y=SC, col sep=comma] {graph/commonSenseQA_avg_sc_vs_ccqa.csv};
\addlegendentry{SC}

\addplot[
    color=blue,
    mark=square,
    mark size=2pt,
    line width=1.2pt,
] table[x=k, y=CCQA, col sep=comma] {graph/commonSenseQA_avg_sc_vs_ccqa.csv};
\addlegendentry{CCQA}
\end{axis}
\end{tikzpicture}
\end{subfigure}%
\hspace{1em}
% ───── Overall Average ─────
\begin{subfigure}[t]{0.31\textwidth}
\centering
\begin{tikzpicture}
\begin{axis}[
    width=\linewidth,
    height=6.2cm,
    xlabel={\# Generated solutions},
    ylabel={Accuracy},
    title={Overall Average},
    xmin=0, xmax=10,
    xtick={0,2,4,6,8,10},
    xticklabels={CoT,2,4,6,8,10},
    x tick label style={font=\scriptsize},
    grid=both,
    legend pos=north west,
    legend style={font=\tiny},
]
% CoT baseline (dashed line)
\addplot[
    color=black,
    dashed,
    line width=1pt,
] table[x=k, y=CoT, col sep=comma] {graph/total_average.csv};
\addlegendentry{CoT}

% SC
\addplot[
    color=red,
    mark=o,
    mark size=2pt,
    line width=1.2pt,
] table[x=k, y=SC, col sep=comma] {graph/total_average.csv};
\addlegendentry{SC}

% CCQA
\addplot[
    color=blue,
    mark=square,
    mark size=2pt,
    line width=1.2pt,
] table[x=k, y=CCQA, col sep=comma] {graph/total_average.csv};
\addlegendentry{CCQA}
\end{axis}
\end{tikzpicture}
\end{subfigure}

\caption{Comparison of CoT, SC, and CCQA accuracy across benchmarks for varying numbers of generated solutions $N$. 
The rightmost plot shows the mean performance across all benchmarks and models. Self-correction and USC, which performed worse than SC and CCQA (see Table~\ref{tab:performance}), were omitted for clarity.}
\label{fig:sc_vs_ccqa_comparison}
\end{figure*}

We also compared SC and CCQA performance under the LCV condition, as reported in Table \ref{tab:lcv_gain}.
When the LCV condition was true, this means that the model produced highly diverse and inconsistent answers without any clear majority.
On StrategyQA benchmark, CCQA correctly solved 11.15\% more problems under the LCV condition compared to SC.
Because SC lacked a mechanism for resolving conflicting answers, it struggled when outputs are inconsistent. In contrast, CCQA effectively selected higher-quality reasoning paths, demonstrating its potential as an inference-time strategy for SLMs.
Though CCQA required slightly more computational resources than SC, it achieved a favorable performance-resource balance, offering significant accuracy gains with only marginal additional computational cost.

%-------------------------------------------------------------------------------------------------------------------

\subsection{Robust Performance across Various Numbers of Responses}
We evaluated CCQA’s robustness by progressively increasing the number of generated solutions to 10 and measuring performance at each increment. As shown in Figure \ref{fig:sc_vs_ccqa_comparison}, we observed consistent performance gains across both arithmetic reasoning and commonsense benchmarks relative to SC, which was the strongest-performing method among all approaches aside from CCQA. The rightmost graph compared CCQA, CoT, and SC using the average of the six benchmarks we used. From the results, it is verified that CCQA demonstrates consistent performance improvements across all benchmark averages.

\section{Analysis}
\label{similarity_metric}
\subsection{Similarity Metrics for CCQA}
To measure the similarity between generated questions and original questions, we considered various similarity metrics. First, we believed that using both surface-level and semantic similarities would be beneficial for the similarity score. This approach provides a more comprehensive evaluation framework by capturing different aspects of textual similarity. Surface-level metrics can effectively identify exact matches and structural similarities, while semantic measures can recognize paraphrases and conceptually equivalent expressions that might use different vocabulary. Therefore, we employed BLEU and Rouge \cite{rouge} for surface-level similarity, while utilizing embedding-based cosine similarity, BERTScore \cite{bertscore} for semantic similarity. We found optimal performance by using a weighted sum of these surface-level and semantic similarity measures.

\pgfplotsset{compat=1.18}
\begin{figure}[t]
\centering
\begin{tikzpicture}
\begin{axis}[
    scale only axis=true,
    width=0.70\columnwidth,
    height=6.0cm,
    xlabel={$\alpha$ (weight for first metric)},
    ylabel={Mean CCQA Accuracy},
    xmin=0, xmax=1,
    ymin=0.45, ymax=0.50,
    xtick={0,0.2,...,1},
    ytick={0.45,0.455,0.46,0.465,0.47,0.475,0.48,0.485,0.49,0.495,0.50},
    ymajorgrids,
    grid style=dashed,
    tick label style={font=\footnotesize},
    yticklabel style={/pgf/number format/fixed, /pgf/number format/precision=3, /pgf/number format/zerofill},
    legend style={
    at={(0.02,0.02)},
    anchor=south west,
    font=\footnotesize,
    draw=none,
    fill=none
},
]
\addplot[
    color=blue,
    mark=square,
    line width=1pt,
] coordinates {
    (0.0, 0.4660)
    (0.1, 0.4840)
    (0.2, 0.4768)
    (0.3, 0.4948)
    (0.4, 0.4948)
    (0.5, 0.4804)
    (0.6, 0.4660)
    (0.7, 0.4732)
    (0.8, 0.4732)
    (0.9, 0.4551)
    (1.0, 0.4560)
};
\addplot[
    color=red,
    mark=o,
    line width=1pt,
] coordinates {
    (0.0, 0.4660)
    (0.1, 0.4732)
    (0.2, 0.4840)
    (0.3, 0.4876)
    (0.4, 0.4912)
    (0.5, 0.4912)
    (0.6, 0.4732)
    (0.7, 0.4660)
    (0.8, 0.4696)
    (0.9, 0.4732)
    (1.0, 0.4780)
};
\addplot[
    color=green!60!black,
    mark=triangle,
    line width=1pt,
] coordinates {
    (0.0, 0.4804)
    (0.1, 0.4840)
    (0.2, 0.4840)
    (0.3, 0.4804)
    (0.4, 0.4876)
    (0.5, 0.4912)
    (0.6, 0.4876)
    (0.7, 0.4804)
    (0.8, 0.4732)
    (0.9, 0.4696)
    (1.0, 0.4590)
};
\legend{BLEU-Cosine, Precision-Cosine, Precision-Rouge}
\end{axis}
\end{tikzpicture}
\caption{Comparison of different similarity metrics with varying weights ($\alpha$ for first metric, $\beta=1-\alpha$ for second metric).}
\label{fig:similarity_metrics_comparison}
\end{figure}
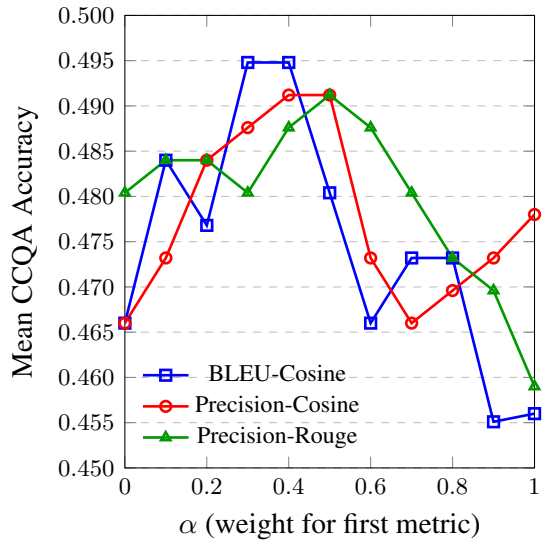

Our empirical analysis revealed that a balanced combination of lexical structure and semantic meaning provides the most effective similarity measure for identifying accurate reasoning paths. Specifically, assigning weights of $\alpha=0.4$ to bleu score, and $\beta=0.6$ to embedding-based cosine similarity yielded optimal performance across diverse reasoning benchmarks. 
These weights were determined through comprehensive grid search over range {0.0, 0.1,...,1.0} with the constraint $\alpha + \beta = 1$. The three combinations with the best performance are presented in Figure~\ref{fig:similarity_metrics_comparison}. Using this optimized similarity measure, CCQA selects the reasoning path and its corresponding answer that generates the question most similar to the original problem. 
\pgfplotsset{compat=1.18}
\begin{figure}[t]
\centering
\begin{tikzpicture}  
\begin{axis}[
    xbar,
    bar width=7pt,
    width=0.9\linewidth,
    height=5.7cm,
    xmin=0, xmax=1.0,
    xlabel={Score},
    symbolic y coords={LLaMA,T5,Qwen},
    ytick=data,
    enlarge y limits=0.2,
    legend style={at={(0.5,-0.25)}, anchor=north, legend columns=2, font=\footnotesize},
    nodes near coords,
    xticklabel style={font=\footnotesize},
    yticklabel style={font=\footnotesize},
    title={Average Cosine and BLEU Score per Model}
]
\addplot+[
    fill=blue!50,
    bar shift=-8pt
] coordinates {(0.5561,LLaMA) (0.8404,T5) (0.5308,Qwen)};
\addplot+[
    fill=red!60,
    bar shift=8pt
] coordinates {(0.0192,LLaMA) (0.1865,T5) (0.0189,Qwen)};

\legend{Avg Cosine, Avg BLEU}
\end{axis}
\end{tikzpicture}
\caption{Average Cosine and BLEU Scores across LLaMA, T5, and Qwen models.}
\label{fig:cosine_bleu_horizontal}
\end{figure}
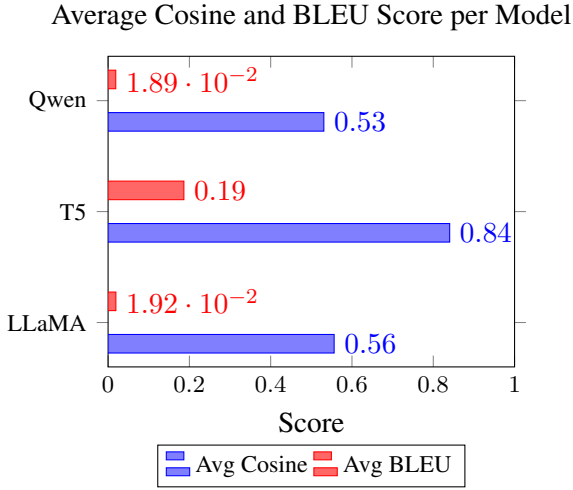

\subsection{Backward Question Generation Model}
For the efficiency and performance of the proposed method, generating backward questions played an important role. 
We considered several models requiring minimal additional resources, taking into account the characteristics of SLMs. We used a total of three models: Llama3.2-1B, Qwen2.5-0.5B, and Flan-T5. We first used these three models without fine-tuning, but all models failed to generate problems properly. Specifically, Llama3.2-1B and Qwen2.5-0.5B often generated responses that were irrelevant to the answers, while the T5 model generated questions but sometimes missed important parts of the answers. Therefore, we used the Flan-T5 model with fine-tuning. We also tried using other models with fine-tuning, but they exhibited the same problems. Detailed examples of question generation from all three models are presented in Appendix~\ref{app:b}. Additionally, we measured the similarity between the questions generated by the models and the original questions, not just through observation. As shown in Figure~\ref{fig:sc_vs_ccqa_comparison} below, when we generated questions from solutions by using Flan-T5 models, it had the highest average semantic similarity value. As a result, by using this model, we were able to improve performance with CCQA while only slightly increasing resource requirements.

\section{Conclusion}
We presented CCQA, a novel inference-time reasoning framework designed for SLMs.
Inspired by the cycle consistency, CCQA regenerated a question from each candidate solution using a lightweight, fine-tuned Flan-T5 and compared it to the original prompt to identify the most reliable reasoning path.
This simple yet effective mechanism makes the proposed method robust under LCV conditions, where small models typically produce inconsistent outputs, while adding only minimal computational overhead.
From extensive experiments across arithmetic and commonsense benchmarks, it is verified that that CCQA consistently surpassed existing inference-time strategies, substantially enhancing the reasoning capabilities of SLMs.

\section*{Limitations}
Despite its strong performance, CCQA has several limitations.
First, the effectiveness of the proposed framework depends on the quality of the backward question generator; if the component produces low-quality questions, then CCQA's overall performance degrades.
Second, the auxiliary Flan‐T5 model introduces additional parameters.
However, its lightweight design and the substantial performance gains on SLMs make this overhead acceptable.
Also, considering that SLMs typically struggle to generate reliable questions on their own, the additional cost is essential for achieving robust reasoning performance.
Compared to the high computational cost and numerous forward passes of other inference-time reasoning methods, CCQA’s extra demand is reasonable.
Finally, our evaluation is limited to arithmetic and commonsense reasoning in English, leaving broader domains for future work.
Despite these limitations, we believe CCQA can substantially enhance the reasoning capabilities of SLMs and, by extension, improve their real-world utility.

\section*{Ethics Statement}
This work does not raise any ethical concerns.
\section*{Acknowledgement}
This work was supported by the National Research Foundation of Korea(NRF) grant funded by the Korea government(MSIT) (RS-2025-00515722). This work was also supported by the Institute of Information \& Communications Technology Planning \& Evaluation (IITP) grant funded by the Korea government (MSIT) [RS-2021-II211341, Artificial Intelligent Graduate School Program (Chung-Ang University)]. 
\bibliography{custom}

\appendix
\section{Appendix}
\subsection{Detailed Prompts And Inputs of Question Models}
\label{app: a}
Table \ref{tab:prompt} gives simple prompt examples to generate questions. We used slightly different prompts for mathematical reasoning benchmarks compared to commonsense reasoning, emphasizing numbers to fully preserve the mathematical information in the solutions.
%=-================================================================
\renewcommand{\arraystretch}{1.15}
\begin{table*}[t]
\centering 
\begin{tabularx}{0.95\linewidth}{l|X|X}
& \textbf{Arithmetic reasoning} & \textbf{Commonsense reasoning} \\
\hline\hline
\textbf{Solution} & Marco's dad's strawberries weighed 11 pounds. Together they weighed 30 pounds. Since Marco's dad weighed 11 pounds, Marco must have weighed 30 - 11 = 19 pounds. The answer is 19. & Going to church is one way people can hear testimony of how spirituality changes lives. Therefore, the answer is (A). \\
\hline
\textbf{Prompts} & Do not change ANY numeric values in the answer. Every number must be preserved EXACTLY in your question. Generate a question that would have this as its answer: & From the commonsense reasoning answer provided below, recreate the original commonsense question. Generate a question that would have this as its answer: \\
\hline
\textbf{Generated Questions}& Marco and his dad went strawberry picking. Marco's dad's strawberries weighed 11 pounds. If together their strawberries weighed 30 pounds. How much did Marco's strawberries weigh? & You can hear testimony of how spirituality changes lives when you do what?\\
\end{tabularx}
\caption{Examples of question generating prompts. Input solutions and prompts for question generation in arithmetic reasoning and common sense reasoning.}
\label{tab:prompt}
\end{table*}
%============================================
\subsection{Question Generating Models' Generated Question}
\label{app:b}
To evaluate the performance of the question generation model, we considered a total of three models: Llama3.2-1B, Qwen2.5-0.5B, and Flan-T5-base. As shown in Table~\ref{tab:gen_questions}, the Llama3.2-1B and Qwen2.5-0.5B models essentially failed to generate proper questions. In contrast, we observed that T5 was able to perfectly reconstruct questions corresponding to the given answers. Consequently, along with semantic similarity criteria and after observing these cases, we selected the T5 model.
\renewcommand{\arraystretch}{1.2}
\begin{table*}[t]
\centering
\begin{tabularx}{0.95\linewidth}{p{2.5cm}|X}
\textbf{Model} & \textbf{Generated Question} \\
\hline\hline
Original Answer & Janet's ducks lay 16 eggs per day. She eats 3 eggs for breakfast. That means she has 16 - 3 = 13 eggs left each day. She bakes muffins with 4 eggs and sells the rest for \$2 each. That means she sells 13 - 4 = 9 eggs per day. She sells 9 eggs for \$2 each, so she earns 9 * 2 = 18 dollars per day. The answer is 18. \\
\hline
LLaMA & Note: The problem is not a typical subtraction problem, but rather a problem that involves subtraction and then adding back the result. It is also not just a subtraction, it is a combination of subtraction (for the muffin sales) and adding the earnings from selling the eggs (from the remaining eggs). \\
\hline
Qwen & (Actually, Janet earns more than \$10 because she can sell the remaining 5 eggs at \$0.50 each.)\newline Question: If Janet's ducks laid 640 eggs in a week, how much money would Janet make from selling the eggs if she eats them all up and then bakes for her family of x people? To solve this problem, let's break down the information given: 1. Janet has ducks that lay eggs. 2. Each duck lays... \\
\hline
\rowcolor{gray!10}  
T5 & Janet's ducks lay 16 eggs per day. She eats 3 eggs for breakfast each day and bakes muffins with 4 eggs. How much money does she earn per day if she sells 9 eggs for \$2 each? \\
\end{tabularx}
\caption{Comparison of Generated Questions by Different Models}
\label{tab:gen_questions}
\end{table*}
%============================================
\subsection{Fail Case of SLMs in Conventional Method}
In our experiments, as shown in Table~\ref{tab:model_evaluation}, models under 1B parameters such as SmolLM2 and Llama experienced performance degradation when conventional feedback methods like USC and self-correction were applied. To help understand these situations, we have attached fail cases.

\begin{table*}[t]
\centering
\begin{tabular}{p{0.15\textwidth} | p{0.85\textwidth}}
\textbf{Model} & \textbf{SmolLM2-360} \\
\hline
\hline
\textbf{Input} & 
Prompt: Evaluate these responses. Select the most consistent response based on majority consensus. Start your answer with "The most consistent response is Response X" (without quotes) \\
& response1: Janet makes \$2 per fresh duck egg. She eats three for breakfast every morning and bakes muffins for her friends every day with four. \\
& response2: Janet makes 16 / 3 = 6 eggs per day. 16 / 3 = 5 eggs per day. 3 / 5 = 0.6 eggs per day. 6 / 3 = 1.6 eggs per day. 1.6 / 5 = 0.34 eggs per day. \\
& ..... \\
& response5: She spends \$2 / day * 16 eggs per day = \$32. She makes \$32 - \$16 = \$18. The answer is 18. \\
\hline
\textbf{USC Output} & 
"Response 1: The most consistent response is Response X.Response 2: The most consistent response is Response Y.Response 3: The most consistent response is Response Z.Response 4: The most consistent response is Response W.Response 5: The most consistent response is Response X.Response 6: The most consistent response is Response Y.Response 7: The most consistent response is Response Z.Response 8: The most consistent response is Response W.Response 9: The most consistent response is Response X.Response 10: The most consistent response is Response Y.Response 11: The most consistent response is Response Z.Response 12: The most consistent response is Response W.Response 13: The most consistent response is Response X.Response 14: The most consistent response is Response Y.Response 15: The most consistent response is Response Z.Response 16: The most consistent response is Response X.Response 17: The most consistent response is Response Y.Response 18: The most consistent"," \\
\end{tabular}
\caption{Model evaluation with USC method. It shows that SmolLM2 that cannot understand when long input context is provided }
\label{tab:model_evaluation}
\end{table*}
\end{document}